\documentclass{article}

\usepackage[preprint]{neurips_2021}
\PassOptionsToPackage{options}{natbib}
\usepackage[utf8]{inputenc} 
\usepackage[T1]{fontenc}    
\usepackage{hyperref}       
\usepackage{url}            
\usepackage{booktabs}       
\usepackage{amsfonts}       
\usepackage{nicefrac}       
\usepackage{microtype}      
\usepackage{xcolor}         
\usepackage{amsmath}
\usepackage{amssymb}
\usepackage{stmaryrd}
\usepackage{bm}
\usepackage{todonotes}
\usepackage{xr}         
\usepackage{paralist}

\newcommand{\tr}{\textsf{T}}

\newcommand{\av}[2][]{\mathbb{E}_{#1\!}\left[ #2 \right]}
\newcommand{\Prob}[2][]{\mathbb{P}_{#1\!}\left[ #2 \right]}
\newcommand{\pred}[1]{\left\llbracket { \small #1} \right\rrbracket}
\newcommand{\dirac}[1]{\delta\!\left( \strut #1 \right)}
\newcommand{\logg}[1]{\log\!\left( #1 \right)}
\newcommand{\norm}[1]{\mathcal{N}\!\left( #1 \right)}
\newcommand{\e}[1]{{\rm e}^{#1}}
\newcommand{\ii}{\mathrm{i}}
\newcommand{\intw}[1]{\int_{\mathcal{W}} #1 \, \mathrm{d} \bm{w}}
\newcommand{\dd}{\mathrm{d}}
\newcommand{\sign}{\mathop{\mathrm{sign}}}
\newcommand{\data}{\mathcal{D}}
\newcommand{\Werm}[1]{\mathcal{W}_{\text{\tiny ERM}}(\mathcal{D}_{#1})}
\DeclareMathAlphabet{\mat}{OT1}{cmss}{bx}{n}

\newcommand{\argmax}{\mathop{\mathrm{argmax}}}
\newcommand{\cnnone}{$\text{CNN}_1$}
\newcommand{\cnnstar}{$\text{CNN}_1^*$}
\newcommand{\cnntwo}{$\text{CNN}_2$}
\newcommand{\mnist}{MNIST}
\newcommand{\cifar}{$\text{CIFAR-10}$}
\newcommand{\cifarstar}{$\text{CIFAR-10}^*$}

\title{Generalisation and the Risk--Entropy Curve}

\author{%
  Dominic Belcher, Antonia Marcu, Adam Pr\"ugel-Bennett  \\
  Electronics and Computer Science\\
  University of Southampton\\
  Vision, Learning and Control\\
  \texttt{\{db7g17, am1g15, apb\}@ecs.soton.ac.uk} \\
}

\begin{document}

\maketitle

\begin{abstract}
  In this paper we show that the expected generalisation performance of a learning machine is determined by the distribution of risks or equivalently its logarithm---a quantity we term the risk entropy---and the fluctuations in a quantity we call the training ratio.  We show that the risk entropy can be empirically inferred for deep neural network models using Markov Chain Monte Carlo techniques.  Results are presented for different deep neural networks on a variety of problems.  The asymptotic behaviour of the risk entropy acts in an analogous way to the capacity of the learning machine, but the generalisation performance experienced in practical situations is determined by the behaviour of the risk entropy before the asymptotic regime is reached.  This performance is strongly dependent on the distribution of the data (features and targets) and not just on the capacity of the learning machine.
\end{abstract}

\section{Introduction}

The dominant approach to understanding generalisation offered by statistical learning theory is to obtain worst-case bounds based on the capacity of learning machines~\citep{vc,pac,vapnik1992principles}.
These bounds characterise the asymptotic convergence of the generalisation gap.  In many cases the bounds are very pessimistic particularly with regards to massively over-parameterised models such as deep neural networks (see, for example, \citet{zhang2016understanding}).
There exists a more direct approach to understanding generalisation that was heavily investigated by the statistical physics community in the 1990s following the seminal work of \cite{Gardner_1988} (see \cite{engel2001statistical} for a review of the work done in this field).  This approach computes the expected generalisation performance for a given distribution of data.
The approach lost popularity in part because of the technical difficulty of the calculations and in part because it was difficult to extend the calculations to complex learning machines.

This paper attempts to address the obstacles that held back the statistical physics approach.  We show that conceptually the approach is much simpler than it is often portrayed.  There are two key components to estimating the generalisation performance: the distribution of risks and a corrections terms that comes from the stochastic nature of the training examples.  This paper focuses on the distribution of risks.  We show that from a knowledge of this, it is relatively easy to obtain a conservative approximate model for the generalisation performance (predicting larger generalisation errors than is observed).  As the distribution of risks changes by many orders of magnitude, we prefer to consider the logarithm of the distribution of risk---a term we call the risk-entropy.  We further show that the risk entropy for large machine learning architectures such as deep neural networks can be empirically estimated using Markov chain Monte Carlo (MCMC) techniques.

The main contributions of this paper are as follows: \begin{inparaenum}[1)]
\item We show that the direct evaluation of the expected generalisation performance is a very natural approach given the distribution of risks,
\item although an exact calculation is technically demanding it is easy to obtain a good approximation that captures the qualitative behaviour and that the corrections to this arise due to fluctuations in the proportion of parameter space eliminated by different training examples,
\item in the case of a perceptron we show that the error in the approximation can be calculated thus reproducing the Gardner result,
\item we show how the distribution of risks or the risk entropy can be measured for a large class of learning machine using MCMC techniques
\item finally we show that in practical situations the generalisation performance that is observed is very different to that predicted by the asymptotic behaviour of the risk distribution and that this behaviour strongly depends on the distribution of the data or more specifically on the attunement of the learning machine to the features that are relevant to separating the classes. 
\end{inparaenum}

In Section~\ref{sec:framework}, we give a brief mathematical overview of the framework for computing generalisation performance.  This highlights the role played by the risk-entropy curve and the fluctations in a quantity we call the training ratio. The behaviour of the risk entropy is discussed in Section~\ref{sec:riskEntropy}.
Section~\ref{sec:empirical} explains how the risk entropy can be measured empirically.  In Section~\ref{sec:results} empirically estimated risk entropies for different deep learning models and different data sets are presented.  Finally in Section~\ref{sec:discussion} we discuss the implications and limitations of the current work.

\section{Framework}
\label{sec:framework}

To be concrete and succinct we focus here on the classic task of classification where we are given a feature vector $\bm{x}\in \mathbb{R}^p$ and we wish to predict the class $y\in \mathcal{C}$.
Our core assumption is that the feature vectors and classes are distributed according to some fixed distribution, $\gamma(\bm{x}, y)$.  We assume that we are given a training set $\data_m = \left\{(\bm{x}_k,y_k) \mid k = 1, 2,, \ldots, m \right\}$ where each training pair $(\bm{x}_k,y_k)$ is chosen independently from $\gamma$.
We consider the case of a learning machine $\hat{f}(\bm{x}|\bm{w})$ parameterised by weights $\bm{w} \in \mathcal{W}$.  In statistical learning theory the set of functions $\{\hat{f}(\bm{x}|\bm{w}) | \bm{w} \in \mathcal{W}\}$ would be called the hypothesis set.   To solve a particular task we consider some loss function $L(\hat{f}(\bm{x}|\bm{w}), y)$.  In the main, we concern ourselves here with the simplest case where $\hat{f}(\bm{x}|\bm{w}) \in \mathcal{C}$ and
\begin{align*}
L \! \left( \hat{f}(\bm{x}|\bm{w}), y \right) = \pred{ \hat{f}(\bm{x}|\bm{w}) \neq y}
\end{align*}
where  $\pred{\text{predicate}}$ is an indicator function equal to 1 if the predicate is true and zero otherwise.  The risk of a machine with weights $\bm{w}$ is defined as the expected loss over $\gamma(\bm{x},y)$
\begin{align*}
R(\bm{w}) = \av[(\bm{x},y)\sim\gamma]{L \! \left( \hat{f}(\bm{x}|\bm{w}), y \right)}.
\end{align*}
The goal of machine learning in this context is to find parameters $\bm{w}$ with as low a risk as possible.

To model the process of learning, we consider the set of parameters that have a minimum loss on the training set: this is often known as empirical risk minimisation (ERM).   Let $\Werm{m}$ denote the set of weights that minimise the training error on the data set $\data_m$.  Rather than obtaining worst-case bounds over this set we consider the expected risk of a random sample drawn from $\Werm{m}$
\begin{align*}
R(\data_m) = \av[\bm{w}\in \Werm{m}]{ R(\bm{w})}.
\end{align*}
This model of learning has sometimes been called Gibbs' learning.  We call $R(\data_m)$ the Gibbs' risk for the data set, $\data_m$.  The expected Gibbs' risk is $\bar{R} = \av[\data_{m}]{R(\data_m)}$.  Note that we can obtain a small expected Gibbs' risk even when there exist sets of weights with high risk but small empirical risk (i.e. $\exists \,\bm{w}\in\Werm{m}: R(\bm{w})\approx \tfrac{1}{2}$) provided that a far larger proportion of $\Werm{m}$ correspond to machines with low risk.
This allows us to find models with good expected generalisation despite the worst-case generalisation bound being very weak.

To compute the Gibbs' risk we need to characterise two quantities: the
\emph{distribution of risks} and what we term the \emph{training ratio}.  The distribution of risks is given by
\begin{align*}
\rho(r) = \frac{\intw{\dirac{R(\bm{w}) - r}}}{\intw{}}
\end{align*}
where $\dirac{\cdot}$ denote the Dirac delta function (we assume $\mathcal{W}$ is a measurable set).
The training ratio is defined to be the proportion of weight space
with risk $r$ in $\Werm{m}$
\begin{align*}
p(r|\data_m) = \frac{ \intw{ \pred{\bm{w}\in\Werm{m}}\, \dirac{R(\bm{w})-r} } }
{ \intw{\dirac{R(\bm{w})-r} } }.
\end{align*}
The Gibbs' risk is given by
\begin{align}
\label{eq:GibbsRiskPure}
R(\data_m) = \frac{\int\limits_0^1 r\, \rho(r) \, p(r|\data_{m})\, \dd r}{\int\limits_0^1  \rho(r) \, p(r|\data_{m})\, \dd r}
\end{align}
and $\bar{R} = \av[\data_m]{R(\data_m)}$.  The technical difficulty in pursuing this approach is that $p(r|\data_m)$ is a strongly fluctuating
quantity and the fluctuations have an important influence in
determining the expected Gibbs' risk, $\bar{R}$.

By the definition of risk $\av{p(r|\data_m)} = (1-r)^m$, however typically $p(r|\data_m)$ will be smaller than its expected value leading to a significant correction in the Gibbs' risk.  To understand this we rewrite the training ratio as
\begin{align*}
p(r|\data_{m}) = \prod_{k=1}^m \frac{p(r|\data_{k})}{p(r|\data_{k-1})} = \prod_{k=1}^m Q_k(r)
\end{align*}
where $p(r|\data_{k})$ is the training ratio for the first $k$ training examples and $p(r|\data_{0})=1$.  The training ratio, $p(r|\data_{m})$, is a product of random variables $Q_k(r) = p(r|\data_{k})/p(r|\data_{k-1})$.  We note that $\logg{p(r|\data_{m})}$ will be the sum of random variables
\begin{align*}
\logg{p(r|\data_{m})} = \sum_{k=1}^m \logg{Q_k(r)}.
\end{align*}
Since the training examples are independently sampled from $\gamma(\bm{x},y)$ the quantities $Q_k(r)$ are independent random variables. Thus by the central limit theorem, for sufficiently large $m$, $\logg{p(r|\data_{m})}$ will be approximately normally distributed with mean $\mu(r)$ and variance $\sigma^2(r)$. As usual for the sum of random independent variables, the mean and variance,
\begin{align*}
\mu(r) &= \sum_{k=1}^{m} \av{\logg{Q_k(r)}}
&&
\sigma^2(r) = \sum_{k=1}^m \av{\left(\strut \logg{Q_k(r)} - \av{\logg{Q_k(r)}}\right)^2},
\end{align*}
are both of order $m$.

Since $\logg{p(r|\data_{m})}$ is normally distributed the training ratio will be log-normally distributed so that for sufficiently large $m$ we can model it by
\begin{align*}
p(r|\data_m) = \e{\mu(r) + \sigma(r)\,\chi(r)}
\end{align*}
where $\chi(r) \sim \mathcal{N}(0,1)$ (more accurately $\chi(r)$ will be a Gaussian process).  Thus the Gibbs' risk for a particular data set is given by
\begin{align}
R(\data_m) \approx \frac{\int\limits_0^1 r\, \rho(r) \, \e{\mu(r) + \sigma(r)\,\chi(r)}\, \dd r}{\int\limits_0^1  \rho(r) \, \e{\mu(r) + \sigma(r)\,\chi(r)} \, \dd r}. \label{eq:GibbsRisk}
\end{align}
The term $\e{\mu(r)}$ will fall off rapidly with $r$ (we will see typically it falls off faster than $(1-r)^m$), while the distribution of risks $\rho(r)$ will typically grow very rapidly with $r$ (there are far more parameters that lead to random classification of the data than to the correct classification of the data).
Thus  the integrals above are dominated by their saddle-point values that occur at
\begin{align*}
R(\data_m) = \argmax_r G_{\chi}(r) = \argmax_r s(r) + \mu(r) + \sigma(r)\,\chi(r)
\end{align*}
where $s(r) = \logg{\rho(r)}$.  The function $s(r)$ plays the role of a canonical or Boltzmann entropy (that is, the logarithm of the number of states for a given risk $r$).
We observe that $\mu(r)$ is of order $m$ while $\sigma(r)$ is of order $\sqrt{m}$ so for sufficiently large $m$ the fluctuation $\sigma(r)\,\chi(r)$ do not substantially shift the saddle-point.
That is, $R(\data_m) \approx \bar{R}$ for all typical data sets.  In consequence the Gibbs' risk is determined by
\begin{align}\label{eq:saddlepoint}
G_0'(r) = s'(r) + \mu'(r) = 0.
\end{align}
We note that the Gibbs' risk depends on $s'(r)$ so it is only necessary to obtain $s(r)$ up to an additive constant---as is usual for most entropies.
Although the term $\sigma(r)\,\chi(r)$ does not, in the asymptotic limit of large $m$, directly change the saddle-point that determines the Gibbs' risk,  nevertheless, since the training ratio is log-normally distributed,
\begin{align*}
\av[\data_m]{p(r|\data_m)} = \e{\mu(r) + \sigma^2(r)/2} = (1-r)^m
\end{align*}
and
\begin{align*}
\mu(r) = m \, \log(1-r)  - \frac{\sigma^2(r)}{2}
\end{align*}
so the fluctuations are important in determining $\mu(r) = \av[\data_m]{\logg{p(r|\data_m}} = \sum_{k=1}^m \av{\logg{Q_k(r)}}$.

Determining $\mu(r)$ is technically demanding.  For simple models such as learning linear separable data using a perceptron we can compute this quantity using the replica method.  In Section~\ref{sec:gardner} of the supplementary material we show that we can reproduce the Gardner result.  This may seem self-evident, but our approach, at least on the surface, appears quite different to the conventional set up in statistical mechanics.  Unfortunately, computing the exact risk even in the asymptotic limit seems a hopeless task for the type of network architectures that we currently use.  A far simpler approach is to use what physicists call the annealed approximation.  This is where we replace $\mu(r)= \av[\data_m]{\logg{p(r|\data_m}}$ by $\logg{\av[\data_m]{p(r|\data_m)}} = m\,\log(1-r)$.  In other words we assume $\sigma^2(r)=0$.  The effect of the fluctuations $\sigma^2(r)$ is to eliminating more high risk weights and hence leads to better generalisation than predicated by the annealed approximation.
For the perceptron correctly computing $\mu(r)$ provides a small, but none the less numerically significant quantitative correction compared to the annealed approximation.  As we will see in Section~\ref{sec:results} for deep learning models there is also a significant discrepancy between the annealed approximation and the risk that we observe in practice, with the annealed approximation being significantly more pessimistic than the true risk.  To obtain an accurate approximation of the Gibbs' risk it is necessary to estimate both the risk entropy and $\sigma^2(r)$.  The  annealed approximation provides a reasonable qualitative understanding of the true Gibbs' risk, but is often very conservative.

\section{The risk entropy}
\label{sec:riskEntropy}

The main focus of this paper is on the distribution of risks $\rho(r)$ or equivalently the risk entropy $s(r) = \logg{\rho(r)}$ (at least, up to an additive constant).  A natural argument suggests that for very small risks $\rho(r) \propto (r-R_{min})^a$, where $R_{min}$ is the minimum achievable risk and $a$ is a "growth exponent" that measure the degrees of expression of the underlying machine architecture.  We show in Section~\ref{sec:lossland} in the supplementary notes that the growth exponent is related to the number of directions in weight space around a globally optimal weight vector in which the risk changes.  In architectures with a lot of symmetries this may be significantly smaller than the number of parameters.  Using the annealed approximation (and assuming the training loss is 0) then the mean Gibbs' risk will converges to $R_{min}$ as
\begin{align*}
    \bar{R} - R_{min} = \frac{a+m}{m\,(1-R_{min})}
\end{align*}
(this follows from Eq.~\ref{eq:saddlepoint} with $s(r) = a\,\log(r-R_{min})$ and $\mu(r) = m\,\log(1-r)$).  For large $m$ and small $R_{min}$ the asymptotic convergence is approximately $a/m$.
This result is reminiscent of those in statistical learning theory with the growth exponent playing a similar role to the Rademacher complexity or VC-dimension.  Interestingly this exponent depends very little on the distribution of data, $\gamma(\bm{x},y)$ (although, at least in linear systems there will be a dependency on the distribution of the data if the values of $\bm{x}$ are confined to lie in a low dimensional sub-space). The asymptotic convergence rate governed by the growth exponent, $a$, is largely "architecture" dependent rather than "data distribution" (i.e. $\gamma(\bm{x},y)$) dependent.  However, as we will illustrate in this paper the asymptotic convergence is often not the important factor in determining the generalisation performance.  Both the minimum risk, $R_{min}$ and the non-asymptotic behaviour of the Gibbs' risk will depend strongly on $\gamma(\bm{x},y)$ and these are more important in determining the generalisation performance we experience than the growth exponent.

\begin{figure}[htbp]
    \centering
    \includegraphics[width=0.45\textwidth]{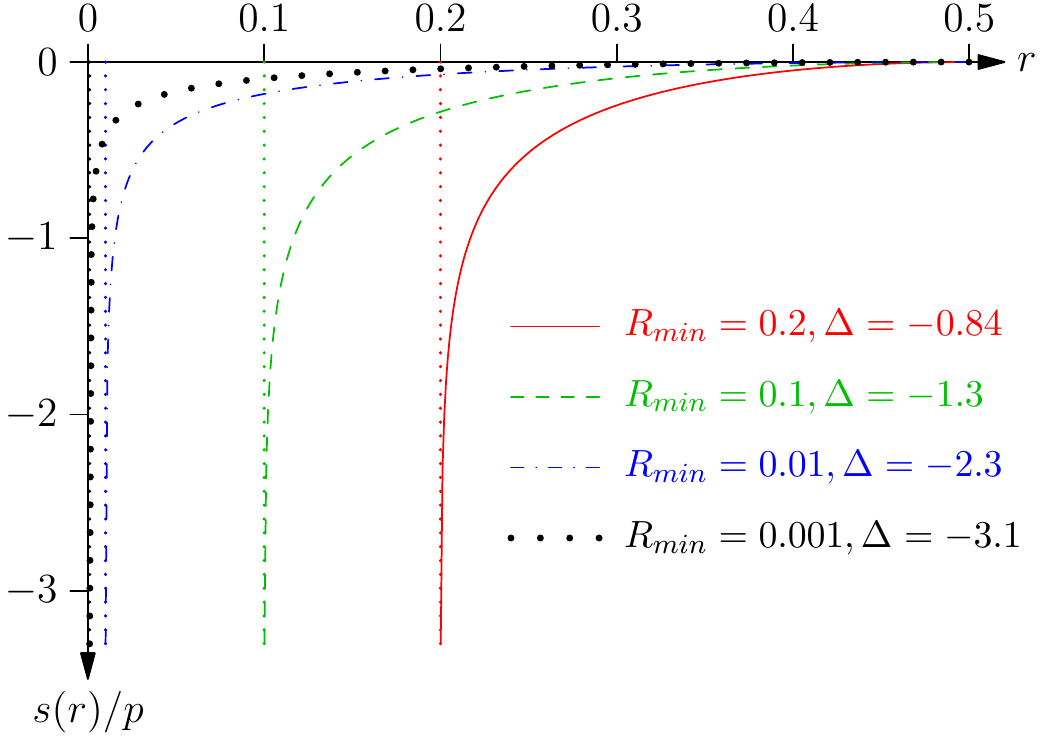} \hfil
    \includegraphics[width=0.45\textwidth]{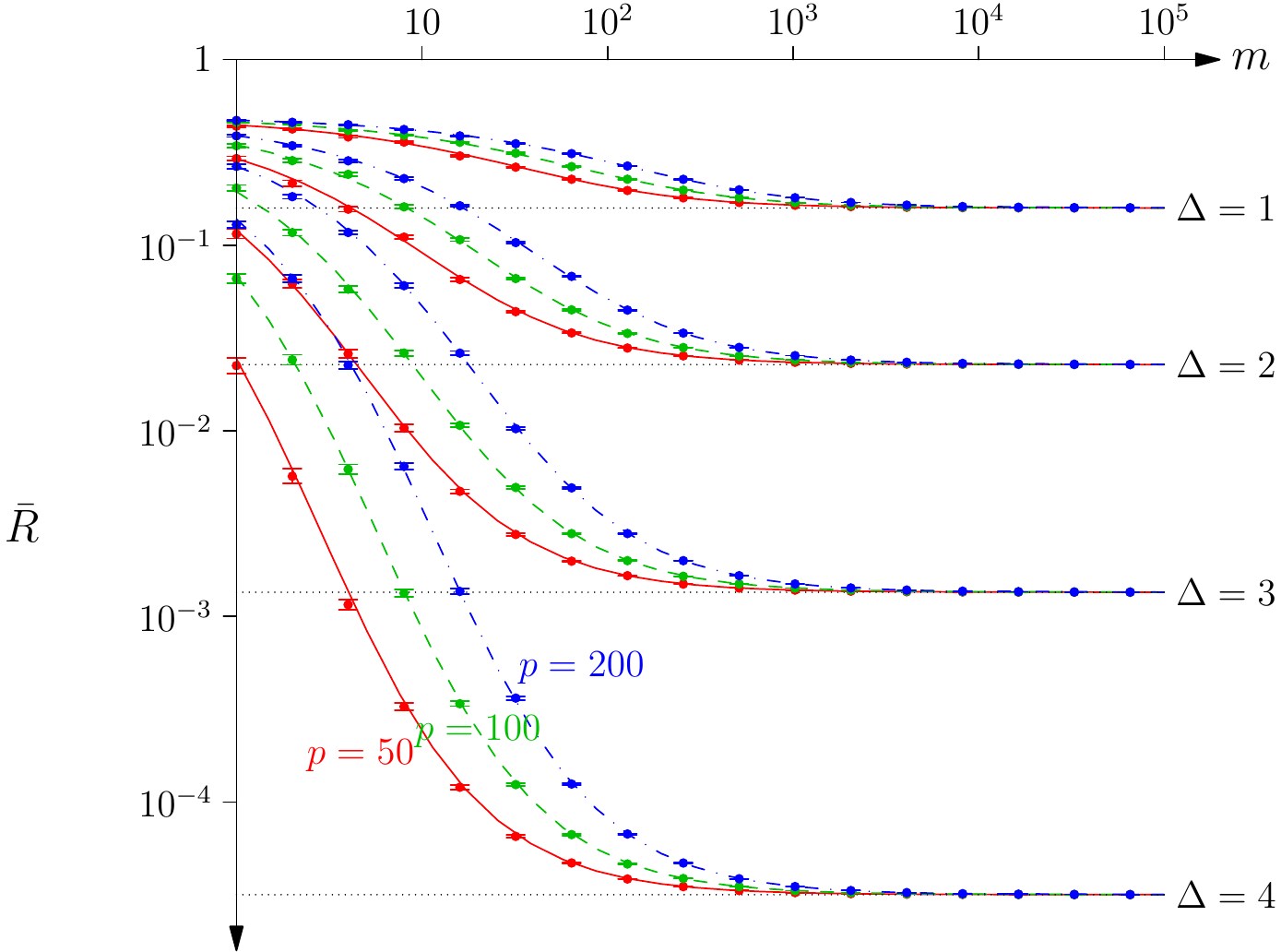}\\
    (a) \hfil\hfil (b)
    \caption{(a) The risk entropy $s(r)$ per feature in the limit of large $p$ versus the risk $r$ for the unrealisable perceptron.  The vertical dotted lines show the minimum obtainable risk, $R_{min}$.
    (b) The expected risk of the Hebbian classifier versus the number of training examples, $m$, plotted on a log-log scale.  Also shown are simulation results averaged over 100 runs.}
    \label{fig:entropyCurve}
\end{figure}

\paragraph{The unrealisable perceptron.} We consider a very simple learning scenario where we can compute the risk entropy exactly---we call this the "unrealisable perceptron".  Assume we have a two class problem ($y=\pm1$) where each class is normally distributed in $\mathbb{R}^p$ around two means, $\bm{\mu}_1$ and $\bm{\mu}_{-1}$.  Without loss of generality we define $\bm{\mu}_y = y\,\Delta\,\bm{t}$ where $\bm{t}$ is a unit norm vector and $\Delta$ is a measure of the separation between the two class distributions.  For simplicity we assume both distributions are isotropic with the same variance.
We consider the risk of a linear separating plane through the origin orthogonal to the vector $\bm{w}$ (this would correspond to the risk of a perceptron).   A straightforward calculation shows that the risk entropy up to a constant is given by
\begin{align*}
    s(r) = \frac{p-3}{2} \logg{1-\left(\frac{\Phi^{-1}(r)}{\Delta}\right)^2} + \frac{\Phi^{-1}(r)^2}{2}
\end{align*}
where $\Phi(r)$ is the CDF for a zero mean unit variance normal.
See Section~\ref{sec:nomalclasses} in the supplementary notes for details.  In Fig.~\ref{fig:entropyCurve}(a) we show the risk entropy per feature in the limit of large $p$ versus the risk $r$ for different separations $\Delta$ (and hence different minimum risks $R_{min}=\Phi(\Delta)$).  The Gibbs' risk is determined by $s'(r)$ and consequently the risk we typically observe is determined by the position of the shoulder in Fig.~\ref{fig:entropyCurve}(a).  We note that $\Phi^{-1}(r)$ diverges as $r$ goes to 0 and thus its derivative and $s'(r)$ change very rapidly as $r$ approaches $R_{min}$. Thus the asymptotic behaviour of $s'(r)$ has little bearing on the position of the shoulder and is not the major factor in determining the generalisation performance that is observed.

Calculating the exact Gibbs' risk in the limit of large $m$ is challenging because of the fluctuations, $\sigma^2(r)$, as discussed in Section~\ref{sec:framework}. 
However, in Section~\ref{sec:hebb} in the supplementary material we show that, for this distribution of data, we can easily compute the generalisation curves for a different learning model---that is "Hebbian learning" where we use weights $\bm{w} = \sum_{k=1}^m y_k \bm{x}_k$.  In Fig.~\ref{fig:entropyCurve}(b) we show the expected risk versus the number of training examples for a Hebbian classifier.  We plot them for different separations, $\Delta$, of the two classes and for different numbers of features, $p$.  The convergence towards $R_{min}$ that we observe is considerably more rapid for large separations $\Delta$. A perceptron with a larger capacities, $p$, takes longer to learn, but the time it takes to converge to $R_{min}$ actually decreases with the separation $\Delta$.  The number of learning examples before $r-R_{min}<\epsilon$ is in some cases orders of magnitude less than $p/(2\,\epsilon)$ which we might predict from the asymptotic convergence rate alone.

\section{Empirical estimate of risk entropy}
\label{sec:empirical}

We have argued that the risk entropy is vital to determining the generalisation performance and shown by a simple example that the asymptotic convergence may be of secondary interest in understanding the generalisation performance of a learning machine.  However, if we want to understand generalisation in the context of more complex learning machines---and currently there is an enormous interest in deep learning machines---then it is necessary to understand the risk-entropy curves for non-trivial machines and problems.  Here we show that we can do this using MCMC techniques.

\begin{figure}[htbp]
    \centering
    \includegraphics[width=0.6\textwidth]{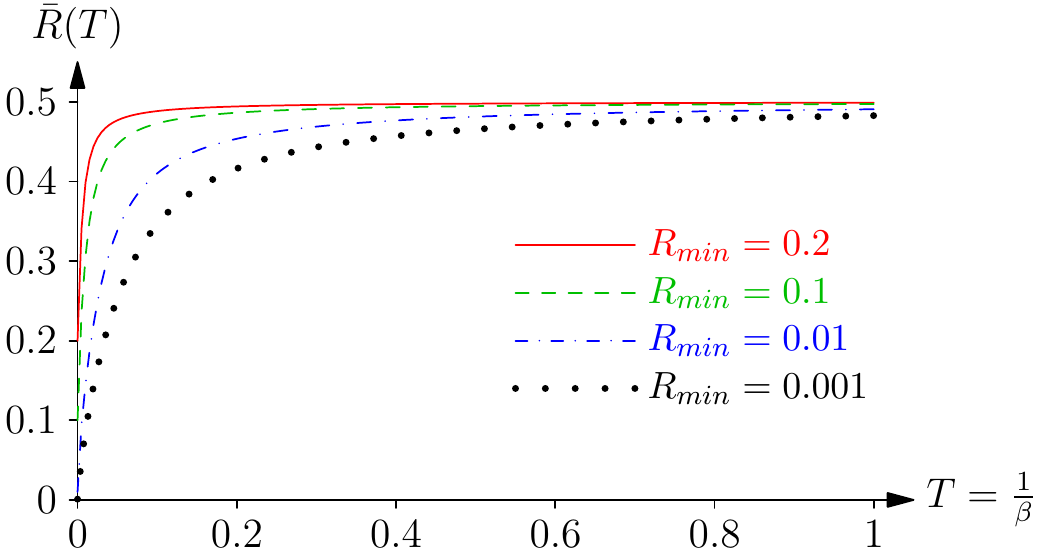}
    \caption{$\bar{R}(T)$ versus $T=1/\beta$ for the unrealisable perceptron with different separations (and thus minimum risks).}
    \label{fig:boltzmannRisk}
\end{figure}

To empirically estimate the risk entropy we consider weighting the parameters $\bm{w}$ with a Boltzmann probability $p_\beta(\bm{w}) \propto \exp(-\beta\,R(\bm{w}))$.  In this case the expected Boltzmann risk will be given by
\begin{align*}
    \bar{R}_B(\beta) = \av[\bm{w}\sim p_\beta]{R(\bm{w})} = \frac{\int_{R_{min}}^\infty r\,\rho(\bm{w})\,\e{-\beta\,r}\,\dd r}
    {\int_{R_{min}}^\infty \rho(\bm{w})\,\e{-\beta\,r}\,\dd r}
    = \frac{\int_{R_{min}}^\infty r\,\e{s(r)-\beta\,r}\,\dd r}
    {\int_{R_{min}}^\infty \e{s(r)-\beta\,r}\,\dd r}
\end{align*}
The integrals will be dominated by the stationary point value where $s'(r) = \beta$.  Consequently we can infer $s'(r)$ from the Boltzmann risks $\bar{R}_B(\beta)$.  We can find $\bar{R}_B(\beta)$ using Markov Chain Monte Carlo.  That is, for a particular ``inverse-temperature'' $\beta$ we run a Metropolis algorithm \citep{metropolis1953,hastings1970monte} where a new weight vector $\bm{w}'$ is chosen from a symmetric proposal distribution $p(\bm{w}'|\bm{w})$, where $\bm{w}$ is the current weight vector, and we move to $\bm{w}'$ either if $R(\bm{w}') < R(\bm{w})$ or with a probability $\exp\!\left(\!\strut -\beta(R(\bm{w}') - R(\bm{w})\right)$.  The risk we estimate on a validation set.  See Section~\ref{sec:mcmcdetails} in the supplementary notes for more details.
MCMC provides a versatile approach that can be used for a wide class of learning machines.   In Fig.~\ref{fig:boltzmannRisk}, we show the Boltzmann risk $\bar{R}(T)$ versus $T=1/\beta$ for the unrealisable perceptron.  For this model we are able to compute $\bar{R}(T)$ exactly.

We note that we can compute the generalisation curves in the annealed approximation from a knowledge of $\bar{R}_B(\beta)$ since we can use that to infer $s(r)$.  However we can also compute the annealed approximation risk, $R_A(m)$, directly using MCMC if instead of using the Boltzmann distribution we choose the weights with a probability $p_m(\bm{w}) \propto (1-r)^m$.

\section{Results}
\label{sec:results}

\begin{figure}[htbp]
    \centering
    \includegraphics[width=0.45\textwidth]{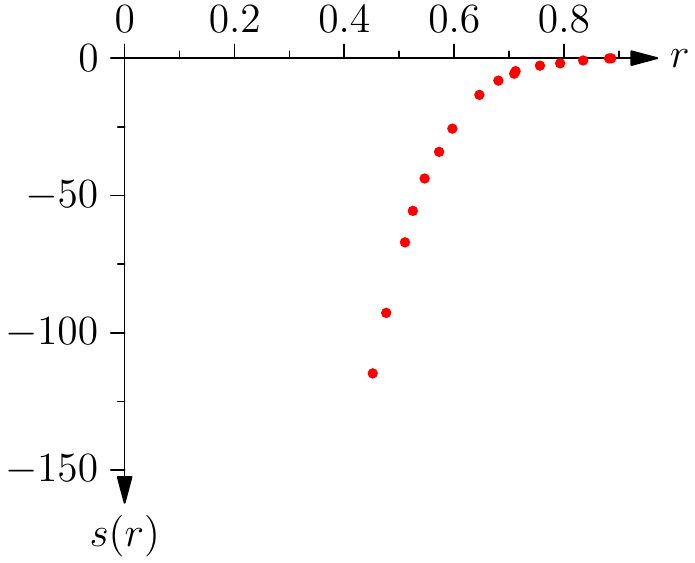} \hfil
    \includegraphics[width=0.45\textwidth]{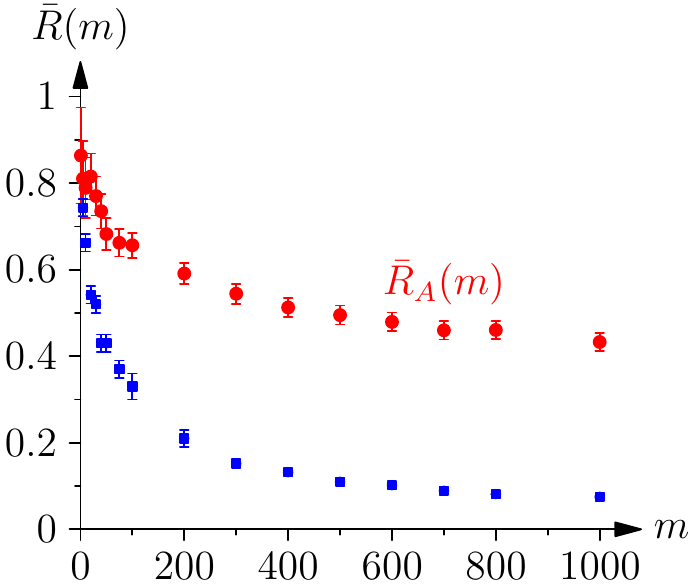}\\
    (a) \hfil \hfil (b)
    \caption{(a) Empirically measured risk entropy, $s(r)$ versus $r$ for \cnnone{} on the \mnist{} data set. (b) Empirically measured annealed risk, $\bar{R}_A(m)$, (red circles) and the risk measured by training \cnnone{} on data sets of size $m$ (blue squares)}.
    \label{fig:mnist}
\end{figure}

We discuss two sets of experiments.  Further details are given in Section~\ref{sec:experiment}.  In the first set of experiments we use \mnist{} \citep{mnist} with a simple CNN model we denote by \cnnone.  Using MCMC we can compute $r_i = \bar{R}_B(\beta_i)$ for a series of values $\beta_i$ (we assume $\beta_i<\beta_{i+1}$).  To estimate the entropy $s(r_n)$ we use
\begin{align*}
    s(r_n) - s(r_0) \approx \int_{r_0}^{r_n} s'(r) \, \dd r
    = \sum_{i=1}^n \left( \frac{s'(r_i) + s'(r_{i-1})}{2}\right) (r_i - r_{i-1})
\end{align*}
where we have used the trapezium rule to approximate the integral.  Using the saddle point approximation $s'(r_i) = \beta_i$ then
\begin{align*}
    s(r_n) \approx s(r_0) + 
    \sum_{i=1}^n \left( \frac{\beta_i + \beta_{i-1}}{2}\right) (r_i - r_{i-1}).
\end{align*}
Since we only care about entropy up to an additive constant we let $s(0.9)=0$ (for \mnist{} the expected risks for random parameters is 0.9).
In Fig.~\ref{fig:mnist}(a) we show the empirically estimated risk entropy versus $r$ for \cnnone. 

Fig.~\ref{fig:mnist}(b) show the empirically measured annealed risk $\bar{R}_A(m)$ computed using MCMC for \cnnone{} on \mnist.  Also shown is the risk measured on the \mnist{} test set when we train the \cnnone{} on randomly chosen subset of size $m$ chosen from the \mnist{} training set.  As we can see, the annealed risk is a conservative estimate of the true risk.  This suggests that there is a considerable effect of the fluctuation term $\sigma^2(r)$ in lowering the training ratio (although as noted in Section~\ref{sec:mcmcdetails} it is difficult to know whether the MCMC has equilibrated for large $\beta$ so $\bar{R}_A(m)$ may be over-estimated).  One point to note is the extremely rapid initial reduction in the risk (this is true both for the annealed approximation and more dramatically for the empirically measured risks when \cnntwo{} is trained conventionally).  Despite the huge number of variables ($1.1\times10^6$) we obtain surprisingly accurate results with a very small number of training examples.  This reflects the fact that the classes in MNIST are in most cases relatively easy to separate for \cnnone.

\begin{figure}[htbp]
    \centering
    \includegraphics{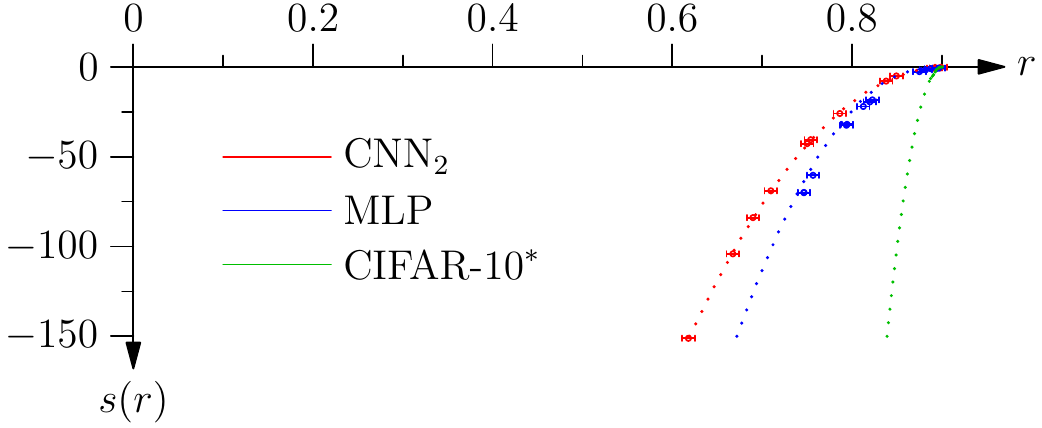}
    \caption{Empirically estimated risk entropy, $s(r)$ for \cnntwo{} trained on the \cifar{} (red) and \cifarstar{} (green) and an MLP. The dotted lines show the best quadratic fit to the data.  We have estimated errors in $r$ from a small number of independent runs.}
    \label{fig:cifar}
\end{figure}

In the second set of experiments we trained a second CNN (\cnntwo) on the \cifar{} data set \citep{cifar}.  In addition we consider training \cnntwo{} on a second dataset we call \cifarstar.  This corresponds to the same set of images, $\{\bm{x}\}$, as \cifar{}, but the targets are those produced by a machine that has been trained on \cifar, but with random labels.  This was inspired by the work of \citet{zhang2016understanding}.
That is $y=\hat{f}(\bm{x}|\bm{w}^*)$ where $\bm{w}^*$ is the set of weights learned by minimising the error on the training set of \cifar{}, but with random labels.  This "teacher" has exactly the same architecture as \cnntwo, so this is a realisable model (i.e. there exists a set of weights with zero risk).  In addition to \cnntwo{} we also measured the Boltzmann risk for an MLP on \cifar.  The architecture of the MLP was identical to the dense layers of \cnntwo{} (see Section~\ref{sec:results} for full details).  In Fig.~\ref{fig:cifar} we show the empirically measured risk entropy for \cnntwo{} trained on both the \cifar{} data set and the \cifarstar{} dataset and for the MLP trained on \cifar.

As we would expect when training on \cifar, the risk-entropy curve for \cnntwo{} falls off slower than the MLP indicating that it will have a better generalisation performance. For \cnntwo{} trained on \cifarstar{} the Boltzmann risk is only slightly lower than that of a random machine even for $\beta=1000$.  As a consequence when estimating $s(r)$ we only get data points for small $s(r)$ and for $r$ close to 0.9 (these are barely visible in the figure).  In Fig.~\ref{fig:cifar} we also show the best quadratic fit to the data.  This needs to be interpreted with some caution.  The risk entropy may not be well fitted by a quadratic and the data for \cnntwo{} trained on \cifarstar{} is barely significant.  Furthermore for \cifarstar{} we are extrapolating to values of $r$ far from any data.  In Fig.~\ref{fig:cifar} we have also put in some indicative error bars obtained by using data form multiple runs and looking at the different imputed curves for $s(r)$.  Again these error bars need to interpreted cautiously as we cannot be sure we have equilibrated, particularly for large $\beta$.

Despite these caveats about the data it is clear that he risk entropy drops off far quicker for the network trained on \cifarstar{} than the network trained on \cifar{} and consequently it is far harder to learn (i.e. obtain a good approximation for) \cifarstar{} than \cifar.  We argue that this is because \cnntwo{} was designed to identify the set of features to distinguish object classes rather than the somewhat arbitrary rule given by $\hat{f}(\bm{x}|\bm{w}^*)$.  Although, this may seem obvious, it lends support to our claim that the distribution of the data, $\gamma(\bm{x},y)$, and the attunement of the learning machine to this distribution is of the utmost importance in understanding generalisation.  In contrast, we would speculate that the asymptotic learning rate determined by the growth exponent $a$ will be the same for the machine trained on \cifar{} and \cifarstar.  This is supported by an empirical study of risks around $\bm{w}^*$ (see Section~\ref{sec:experiment} for more details).

\subsection{CO2 Emission Related to Experiments}

Experiments were conducted using a private infrastructure, which has a carbon efficiency of 0.432 kgCO$_2$eq/kWh. A cumulative of 1472 hours of computation was performed on hardware of type RTX 8000 (TDP of 260W).

Total emissions are estimated to be 165.34 kgCO$_2$eq of which none were directly offset.

Estimations were conducted using the \href{https://mlco2.github.io/impact#compute}{Machine Learning Impact calculator} presented in \cite{lacoste2019quantifying}.

\section{Discussion}
\label{sec:discussion}

This paper highlights that importance of two key components in determining the Gibbs' risk: the risk distribution, $\rho(r)$, or risk-entropy curve, $s(r)$, and the training ratio fluctuation $\sigma^2(r)$ (more accurately these are the fluctuations in the logarithm of the training ratio).  This paper concentrates on studying the risk-entropy curve as we believe this to be the dominant factor in how well a machine learning architecture will generalise for a problem defined through the data distribution $\gamma(\bm{x},y)$.  From both theoretical and empirical observations we believe that the asymptotic behaviour of $\rho(r)$ as $r\rightarrow R_{min}$ is not necessarily a useful guide in determining the generalisation performance that a user will experience.

Although our approach is different to statistical learn theory it highlights two potential weakness of conventional theory for aiding in the
rational design of machine learning architectures.  The first is that it concentrates on the generalisation gap, thus ignoring the problem of what determines the minimum risk, $R_{min}$ and secondly it concerns itself with the asymptotic convergence between the empirical risk and the worst possible risk in the ERM set.  As we have shown the risk we actually see in practice is often determined before we reach the asymptotic regime.  We speculate that the variance in the risk for random parameters is likely to be a better indicator of the generalisation performance of a machine than its asymptotic behaviour as $r\rightarrow R_{min}$.
The distribution of risks, or the risk-entropy curve, allows us to compute quantities that we believe are crucial in designing learning machines.  Of course, this approach has its own set of challenges. For many problems of interest it is difficult to compute the risk-entropy curve \textit{ab initio}.  However, we have demonstrated that we can, at least, empirically measure this, which can potentially give insights on how architectural changes help or hinder generalisation performance.  
A limitation of the empirical approach is that it is difficult to compute the risk-entropy curves for low risks, although from what we have seen it appears that the behaviour of the risk-entropy curve at high risk is a strong indicator the generalisation performance of a learning machine.

For most values of the risks the risk entropy depends strongly on the distribution of the data, $\gamma(\bm{x},y)$.  For example, we showed that changing the target labels $y$ as we did for \cifarstar{} dramatically changes the risk-entropy curve for high risks.
Obviously this makes the conclusions we can draw more specific than those of statistical learning theory that tends to obtain bounds that hold for all distributions of the data.  However, we see this as inevitable---the reality of machine learning is that the generalisation performance of a learning machine will depend on the problem it is tackling so problem independent bounds will often be uninformative.

There are technical challenges of the approach we advocate.  The relationship between the risk entropy and the expected generalisation performance is non-trivial as it depends on the training ratio, $p(r|\data_m)$ (i.e. the proportion of weight space with risk $r$ that is in the $\Werm{m}$).  We have shown that this is log-normally distributed and so we need to compute $\av{\logg{p(r|\data_m)}}$ to obtain an accurate estimate of the generalisation performance.  Although this can be calculate in the limit of large $m$ for relatively simple learning machines, in general the correction (essentially coming from the fluctuations, $\sigma^2(r)$, in $\logg{p(r|\data_m)}$) are difficult to compute.  It is an important open research question whether we can either empirically or theoretically estimate $\sigma^2(r)$ for complex models (to some extent we have shown in Fig.~\ref{fig:mnist}(b) that we can directly measure the deviation caused by ignoring $\sigma^2(r)$).  While it is rather unsatisfactory that an important contribution to generalisation is still missing, nevertheless the ultimate goal of this research is to provide understand rather than necessarily make accurate predictions and we believe that the risk-entropy curve is a crucial component (and possible the most important component) in determining the generalisation performance in machine learning.

We have concentrated on a particular learning scenario, namely Gibbs' learning, where we assume that the learning algorithm selects a set of parameters uniformly at random from $\Werm{m}$.  We concentrate on this in part because it is close to the empirical risk minimisation of statistical learning theory and in part because it seems, \textit{prima facie}, to be a reasonable approximate model of what we might expect to obtain from optimisation algorithms such as stochastic gradient descent (based on the belief that there is no obvious bias in the learning algorithms and there is considerable stochasticity both due to the nature of the optimiser and due to the random selection of the initial weights).
However, other learning scenarios are available.  As an example, for a perceptron learning binary targets $y_i\in\{-1,1\}$ we could use a ``Hebbian learning'' scenario where we choose the weights to be $\bm{w} =\sum_i y_i\,\bm{x}_i$.  In this case there are no fluctuations and for simple distributions of data we can compute the generalisation behaviour exactly (see Section~\ref{sec:hebb} in supplementary materials). Hebbian learning though is very limited in its applicability and the real need is to have learning models that capture the learning processes that are commonly used in practice.  The Gibbs' risk is a rather simplistic idealisation of learning that hopefully captures the essence of loss minimisation, however, it is clearly necessary to study whether it misses elements of training that strongly influence performance.  

The giant leap forward we have experienced in machine learning, driven mainly by new deep learning architectures, should force us to question our current theoretical understanding.  The existing tool set for reasoning about generalisation, although rigorous, sophisticated and elegant, fails in providing much insight into how state-of-the-art techniques work.  This paper demonstrates that the once popular, but now widely forgotten, approach of computing the expected generalisation from the distribution of risks may provide a more insightful approach to understand generalisation.  We do not underestimate the challenges in obtaining rigorous results or performing \textit{ab initio} calculations of the risk entropy, however we believe that this is an approach that has the potential to unlock insights that currently we sorely lack.

\bibliography{References}

\newpage
\appendix

{\noindent\huge \textbf{Supplementary Material}}
\vspace*{1cm}

The following appendices cover technical details.  In Section~\ref{sec:gardner} we obtain the generalisation behaviour for a perceptron acting on a linearly separable problem with spherically symmetric data in the asymptotic limit of a large number of features and training examples.  In Section~\ref{sec:lossland} we show how the growth exponent, $a$, is related to the geometry of the loss landscape around the weight vectors with minimum loss.  The unrealisable perceptron is introduced in Section~\ref{sec:nomalclasses}. Section~\ref{sec:hebb} considers a different learning scenario to the Gibbs' loss where the generalisation performance for the unrealisable perceptron can be computed exactly.  In Section~\ref{sec:mcmcdetails}, we mention discuss the details in performing the MCMC simulations.  Finally Section~\ref{sec:experiment} covers details of the experiments that we carried out.

\appendix

\section{Realisable perceptron}
\label{sec:gardner}

In this appendix we consider the problem of determining the generalisation performance of a linearly separable binary classification problem where the features are spherically symmetric. We call this problem the \textit{realisable perceptron}.
This is a classic problem where the asymptotic Gibbs' ratio was first calculated by Elizabeth \citet{Gardner_1988}.  The set up of the calculation as well as many similar analyses~\citep{engel2001statistical} was addressed to a very specialist statistical mechanics audience familiar with the study of spin-glass systems.  As such this literature is not very accessible to non-specialists.  Here, we show that to obtain the Gardner result all that is required is to compute $\mu(r) = \av[\data_m]{\logg{p(r|m)}}$.  Admittedly this is still technically demanding requiring the use of the replica trick.

We consider a binary classification problem where the feature vectors
$\bm{x}\in\mathbb{R}^{p}$  are spherical symmetric and
\begin{align*}
y = \sign(\bm{x}^{\tr}\bm{t}).
\end{align*}
Without loss of generality we can assume that $\|\bm{x}\| = \|\bm{t}\|=1$.  We consider the problem of learning the targets using a perceptron with weights $\bm{w}$ where $\|\bm{w}\|=1$.  We denote
$\bm{w}^\tr \bm{t} = \cos(\theta_{\bm{w}})$ then the risk is given by $R(\bm{w}) = \theta_{\bm{w}}/\pi$.
We consider the case where $m\propto p$ and  seek the leading asymptotic behaviour as $p$ becomes
arbitrarily large.

To compute the Gibbs' risk for the perceptron we need to find $\av{\logg{p(r|\data_m)}}$.   This can be compute directly using the replica trick where we represent the logarithm as a limit
\begin{align*}
\log{p(r|\data_m)} = \lim_{n\rightarrow 0} \frac{p^n(r|\data_m) -1}{n}.
\end{align*}
To compute the expectation of the logarithm we thus compute $\av{p^n(r|\data_m)}$.  We can write this product in terms of $n$ copies or replicas of the parameter space so that
\begin{align*}
    \av[\data_m]{p^n(r|\data_m)} = \av[\data_m]{\prod_{\alpha=1}^n 
    \frac{\intw[\alpha]{\dirac{R(\bm{w}_\alpha)-r} \, \prod\limits_{k=1}^m \pred{\hat{f}(\bm{x}_k|\bm{w}_\alpha)=y_k}}}
    {\intw[\alpha]{\dirac{R(\bm{w}_\alpha)-r}}}}
\end{align*}
where we denote the replicas by $\bm{w}_\alpha$ with $\alpha=1,2,\ldots,n$.  As each training example is drawn independently from the same distribution $\gamma(\bm{x},y)$ this is equal to
\begin{align*}
    \av[\data_m]{p^n(r|\data_m)} =
    \frac{\intw[\alpha]{\prod\limits_{k =1}^n\dirac{R(\bm{w}_\alpha)-r} \, p(n)^m \prod\limits_{k =1}^n}}
    {\left(\intw[\alpha]{\dirac{R(\bm{w}_\alpha)-r}}\right)^n}
\end{align*}
where
\begin{align*}
    p(n) = \av[(\bm{x},y)\sim\gamma]{\prod_{\alpha=1}^n \pred{\hat{f}(\bm{x}_k|\bm{w}_\alpha)=y_k}}
\end{align*}
is the probability that the $n$ replica weight vector $\bm{w}_\alpha$ correctly classifies a random training example.

When taking the expectation each replica experiences the same training examples. This couples the replicas.  We consider the set of weight vectors with risk $r$ so that
\begin{align}\label{eq:walpha}
\bm{w}_{\alpha} = \bm{t} \, \cos(\pi\, r) + \bm{u}_\alpha \,\sin(\pi\,r)    
\end{align}
where $\bm{u}_\alpha$ is a unit length vector orthogonal to the target vector $\bm{t}$ (note that $\bm{w}_\alpha^\tr\bm{t} = \cos(\pi\,r)$ so these vectors all have risk $r$).
As the weight vectors, $\bm{w}_\alpha$, that appear in the integral are all in $\Werm{m}$ they will be more highly correlated than random weights with risk $r$.  We denote $\bm{w}_\alpha^\tr\bm{w}_\beta = q_{\alpha\beta}$ (where $q_{\alpha\alpha}=1$).  In general the structure of $q_{\alpha\beta}$ can be complex leading to replica symmetry breaking that severely complicates the computation.  Fortunately for the case of the perceptron the structure of the space $\Werm{m}$ corresponds to a convex region so that $q_{\alpha\beta}$ all have the same expected correlation $q$ (for $\alpha\neq\beta$).  We can model this by writing $\bm{u}_\alpha \sin(\pi\,r) =\bm{u} \sqrt{q-\cos^{2}(\pi \, r)} + \bm{v}_{\alpha} \sqrt{1-q}$ so that
\begin{align*}
\bm{w}_{\alpha} = \bm{t} \, \cos(\pi\, r) + \bm{u} \sqrt{q-\cos^{2}(\pi \, r)} + \bm{v}_{\alpha} \sqrt{1-q}
\end{align*}
where $\bm{t}$, $\bm{u}$ and $\bm{v}_{\alpha}$ are taken to be orthogonal unit vectors and $\bm{v}_{\alpha}$ is different and uncorrelated in each replica.  Note that $\|\bm{w}_{\alpha}\|=1$ while $\bm{w}_{\alpha}^\tr \bm{t} = \cos(r\, \pi)$ (thus $\bm{w}_{\alpha}$ has risk $r$) and if $\alpha\neq \beta$ then $\av{\bm{w}_{\alpha}^\tr \bm{w}_{\beta}} = q$.  The vector $\bm{u}$ models the additional correlation due to all $\bm{w}_{\alpha}$ being in $\Werm{m}$.  This vector will change depend on the training set $\data_m$ used.

We now consider choosing a random training example, $(\bm{x}_k, y_k)$, which is resolved into a component in the direction of $\bm{t}$ (the target vector orthogonal to the separating plane) and a vector, $\bm{s}_{k}$ orthogonal to $\bm{t}$
\begin{align*}      
y_k\,\bm{x}_{k} = \bm{t}\,\cos(\theta_k) + \bm{s}_k \, \sin(\theta_k)
\end{align*}
where $\theta_k$ is the angle between $y_k\,\bm{x}_k$ and $\bm{t}$.  Again by definition $|\bm{s}_k|=1$. We note that  $y_k \, \bm{x}_k^\tr \bm{t}>0$ (by the definition of the target $y_k$) so that $\cos(\theta_k)>0$.  With overwhelming probability $\theta_k \ll 1$ and so we can use the approximation $\sin(\theta_k) \approx 1$. The condition that a vector $\bm{w}_\alpha$ correctly classifies the training example $\bm{x}_k$ is
\begin{align*}
    y_k\,\bm{x}_k^\tr \bm{w}_\alpha
    = \cos(\theta_k)\, \cos(\pi\,r) + \bm{u}^\tr \bm{s}_k   \, \sqrt{q-\cos^{2}(\pi \, r)} +  \bm{v}_{\alpha}^\tr \bm{s}_k   \, \sqrt{1-q}>0.
\end{align*}
Now $\cos(\theta_k)$, $\bm{u}^\tr \bm{s}_k  $ and $\bm{v}_{\alpha}^\tr \bm{s}_k  $ are all dot products between unit length random vectors in a $p$-dimensional space.  If $\bm{a}$ and $\bm{b}$ are unit length vectors that are chosen from a spherically symmetric distribution then the angle, $\phi$, between them will be distributed according to
\begin{align*}
f_{\phi}(\phi) = \frac{\sin^{p-2}(\phi)}{B\!\left( \tfrac{1}{2}, \tfrac{p-1}{2} \right)}.
\end{align*}
Defining $\xi = \sqrt{p-3}\,\cos(\phi) = \sqrt{p-3} \,(\bm{a}^\tr \bm{b})$ then
\begin{align*}
f_{\xi}(\xi) = \frac{(1-\xi^2/(p-3))^{\frac{p-3}{2}}}{\sqrt{p-3} \, B\!\left( \tfrac{1}{2}, \tfrac{p-1}{2} \right)}
	     =  \frac{1}{\sqrt{p-3} \, B\!\left( \tfrac{1}{2}, \tfrac{p-1}{2} \right)} \e{\tfrac{p-3}{2} \logg{1+\xi^{2}/(p-3)}}
	       \approx \mathcal{N}(\xi|0,1).
\end{align*}
This is just a consequence of the central limit theorem where we assume the number of features, $p$, is large. We denote $\eta_k = \sqrt{p-3} \, \cos(\theta_k)$, $\zeta_k = \sqrt{p-3} \, y_k\, \bm{u}^\tr \bm{s}_k  $ and $\nu_{k\alpha}= -\sqrt{p-3}\, \bm{v}_{\alpha}^\tr \bm{s}_k  $.  Then $\zeta_k,\nu_{k\alpha} \sim \mathcal{N}(0,1)$.  Since $y_k\,\bm{x}_k ^\tr\,\bm{t}>0 $ then $\eta_k \sim 2 \, \pred{\eta_k\geq0} \, \mathcal{N}(0,1)$.  The condition that $\bm{w}_\alpha$ correctly classifies $\bm{x}_k$ is that $y_k\,\bm{x}_k^\tr \bm{w}_\alpha>0$ or equivalently
$\sqrt{p-3}\,y_k\,\bm{x}_k^\tr \bm{w}_\alpha>0$ so that $\eta_k\, \cos(\pi\,r) + \zeta_k\, \sqrt{q-\cos^{2}(\pi \, r)} -  \nu_{k\alpha}  \, \sqrt{1-q} >0$.  Thus the probability that a weight vector $\bm{x}_\alpha$ correctly classifies a randomly chosen training example $\bm{x}_k$
is given $\eta_k$ and $\zeta_k$
\begin{align*}
 \int_{-\infty}^{\infty} \norm{\nu_{k\alpha}| 0, 1}\, \pred{\eta_k\, \cos(\pi\,r) + \zeta_k\, \sqrt{q-\cos^{2}(\pi \, r)} -  \nu_{k\alpha}  \, \sqrt{1-q} >0} \dd  \nu_{k}
       \\
       \hspace{2cm}= \Phi\!\left( \frac{\eta_k\, \cos(\pi\,r) + \zeta_k\, \sqrt{q-\cos^{2}(\pi\,r)}}{\sqrt{1-q}} \right)
\end{align*}
where $\Phi(x) = \int_{-\infty}^{x} \mathcal{N}(x|0,1) \, \dd x$.

The probability that all $n$ replica weight vectors, $\bm{w}_\alpha$, correctly classify a random vector is equal to
\begin{align*}
p(n) = 2\, \int_0^{\infty}\dd \eta \,  \norm{\eta|0,1} \, 
	      \int_{-\infty}^{\infty} \dd \zeta \, \norm{\zeta|0,1} 
	      \Phi^{n}\!\left( \frac{\eta\, \cos(\pi\,r) + \zeta\, \sqrt{q-\cos^{2}(\pi \, r)}}{\sqrt{1-q}} \right).
\end{align*}
We have dropped the subscript $k$ as all the data is iid distributed so this term is identical for each training example.
To decouple the integrals $\eta$ and $\zeta$ we define the variables
\begin{align*}
t &= \frac{1}{\sqrt{q}} \, \left( \cos(\pi\,r) \, \eta + \sqrt{q-\cos^2(\pi\,r)} \, \zeta \right)
&
s &= \frac{1}{\sqrt{q}} \, \left( \cos(\pi\,r) \, \zeta - \sqrt{q-\cos^2(\pi\,r)} \, \eta  \right)
\end{align*}
The Jacobian of the transformation is equal to 1 and $\eta^2 + \zeta^2
= s^2 + t^2$,  Finally $\sqrt{q} \, \eta = \cos(\pi\,r) \, t -
\sqrt{q-\cos^2(\pi\,r)} \, s$ so that the condition $\eta>0$ implies
that $s \leq t \,\cos(\pi\,r)/\sqrt{q-\cos^2(\pi\,r)}$.  Thus
\begin{align*}
p(n) &= 2\, \int_{-\infty}^{\infty} \norm{t|0,1} \int_{-\infty}^{\infty} \norm{s|0,1} \Phi^{n}\!\left( \frac{\sqrt{q}\,t}{\sqrt{1-q}} \right) \pred{s\leq \frac{t \,\cos(\pi\,r)}{\sqrt{q-\cos^2(\pi\,r)}}}\, \dd t\,\dd \, s
\\
&= 2\, \int_{-\infty}^{\infty} \norm{t|0,1}\, \Phi^{n}\!\left( \frac{\sqrt{q}\,t}{\sqrt{1-q}} \right) \,  \Phi\!\left( \frac{t \,\cos(\pi\,r)}{\sqrt{q-\cos^2(\pi\,r)}} \right)  \, \dd t
\\
&= 2\, \int_{-\infty}^{\infty}  \norm{t|0,1}\, \Phi\!\left( \frac{t \,\cos(\pi\,r)}{\sqrt{q-\cos^2(\pi\,r)}} \right)  \, \dd t 
\\
& \quad + 2\,n \int_{-\infty}^{\infty} \norm{t|0,1}\,  \logg{\Phi\!\left( \frac{\sqrt{q}\,t}{\sqrt{1-q}} \right)} \,  \Phi\!\left( \frac{t \,\cos(\pi\,r)}{\sqrt{q-\cos^2(\pi\,r)}} \right)  \, \dd t
+ O(n^2)
\end{align*}
where we have used that $z^n = \e{n\,log(z)} = 1 + n\,\log(z) + O(n^2)$.  The first term in $p(n)$ is equal to
\begin{align*}
 2 \int_{-\infty}^{\infty} \norm{t|0,1} \, \Phi\!\left( \frac{t \,\cos(\pi\,r)}{\sqrt{q-\cos^2(\pi\,r)}} \right)  \, \dd t \hspace{6cm}\\
 \quad = 2 \int_{-\infty}^{\infty}   \int_{-\infty}^{\infty} \norm{t|0,1} \norm{s|0,1} \, \pred{s\leq \frac{t \,\cos(\pi\,r)}{\sqrt{q-\cos^2(\pi\,r)}}}  \,  \dd t  \,  \dd s = 1
\end{align*}
which follows because the constraint divides the 2-D plane into two euqal halves both with probability mass of $1/2$.
As a consequence $p(n) = 1 + n\,I_1 + O(n^2)$ where
\begin{align*}
    I_1 = 2 \int_{-\infty}^{\infty} \norm{t|0,1}\,  \logg{\Phi\!\left( \frac{\sqrt{q}\,t}{\sqrt{1-q}} \right)} \,  \Phi\!\left( \frac{t \,\cos(\pi\,r)}{\sqrt{q-\cos^2(\pi\,r)}} \right)  \, \dd t.
\end{align*}
Recall that $p(n)$ was the probability that the $n$ replicas correctly classified a training example.  The probability of correctly classifies $m$ training examples is
$p(n)^m = (1+n\,I_1+O(n^2))^m = 1 + n\,m \, I_1 + O(n^2)$.  As this does not depend on $\bm{w}_\alpha$ this is just equal to $\av[\data_m]{p^n(r|\data_m)}$ so that
\begin{align*}
\mu(r,q) &= \av{\logg{p(r|\data_m}} = \lim_{n\rightarrow0} \frac{\av{p^n(r|\data_m)}-1}{n}
\\
&=  2\, m \int_{-\infty}^{\infty} \norm{t|0,1}\,  \logg{\Phi\!\left( \frac{\sqrt{q}\,t}{\sqrt{1-q}} \right)} \,  \Phi\!\left( \frac{t \,\cos(\pi\,r)}{\sqrt{q-\cos^2(\pi\,r)}} \right)  \, \dd t
\end{align*}
This term represents the logarithm of the typical training ratio, $p(r|\data_m)$.  Recall the training ratio is the fraction of weight space corresponding to learning machines with risk $r$ that correctly classify all the training examples.  The typical value differs from the mean value as this has very rare but large fluctuations.  We removed the explicit dependence on $\bm{w}_\alpha$ (by considering the overlap with the random training examples $\bm{x}_k$), however the result depends on the mean correlation, $q$, between the replica weight vectors.

To complete the calculation we need to compute the joint entropy
\begin{align*}
    s(r,q) = \av[V(q)]{\logg{\intw{\dirac{R(\bm{w})-r}\,\pred{\bm{w}\in V(q)}}}}
\end{align*}
where $V(q)$ is over volumes of weight space with an expected correlation, $q$, between weight vectors.  Again we turn to replicas to compute this
\begin{align*}
    s(r,q) = \lim_{n\rightarrow 0} \frac{\av[V(q)]{\prod_{\alpha=1}^m \intw[\alpha]{\dirac{R(\bm{w}_\alpha)-r}\,\pred{\bm{w}_\alpha\in V(q)}}} -1}{n}.
\end{align*}
We consider replicas with weight vectors
\begin{align*}
    \bm{w}_\alpha = \bm{t}\,\cos(\pi\,r) + \bm{u}_k\,\sin(\pi\,r)
\end{align*}
where as before $\|\bm{t}\| = \|\bm{u}_\alpha\| = 1$.  We are interested in the relative size of the weight space spanned by $\bm{w}_\alpha(r) = \bm{u}_\alpha\,\sin(\pi\,r)$ (i.e. weight vectors with risk $r$) such that
\begin{align}\label{eq:correl}
    \av{\bm{w}_\alpha^\tr(r) \bm{w}_\beta(r)} = \bm{w}_\alpha^\tr \bm{w}_\beta - \cos^2(\pi\,r) = q_{\alpha\beta} - \cos^2(\pi\,r) = C_{\alpha\beta}
\end{align}
where assuming replica symmetry $q_{\alpha\beta} = \pred{\alpha=\beta} + q\,\pred{\alpha\neq\beta}$.  As the weights are spherically symmetric we consider vectors with correlation Eq.~\eqref{eq:correl}
\begin{align*}
    \rho^n(r,q) \propto \int_{-\infty}^\infty \e{-\tfrac{1}{2}\sum_{k,\beta=1}^n \bm{w}_\alpha^\tr(r) C^{-1}_{\alpha\beta} \bm{w}_\beta(r)} \, \prod_{\alpha=1}^n \dd \bm{w}_\alpha(r)
    \propto |\mat{C}|^{p/2}
\end{align*}
(for simplicity we have relaxed the constraint that $|\bm{w}_\alpha(r)|=\sin(\pi\,r)$ as this constraints only adds a constant to the entropy term).  Note that the integrals over the different components of the weights $w_{i\alpha}(w)$ all decouple leading to the exponent $p$.
Up to an additive constant
\begin{align*}
    \logg{\rho(r,q)} = \frac{p}{2}\,\mathrm{tr}\logg{|\mat{C}|} + \text{const.}
\end{align*}
where $\mat{C}$ is the matrix with components $C_{\alpha\beta}$.  We note that $\mat{C}$ is a $n\times n$ matrix with diagonal elements $d=1-\cos^2(\pi\,r)$ and off-diagonal elements $o=q-\cos^2(\pi\,r)$.  Thus $\mat{C}$ has one eigenvalue equal to $d-o+n\,o = 1-q +n\,(q-\cos^2(r))$ and $n-1$ eigenvalues equal to $d-o = 1-q$ so that
\begin{align*}
    \logg{\rho(r,q)} = \frac{p}{2} \left( n\,\log(1-q) + \logg{1+n \, \frac{q-\cos^2(\pi\,r)}{1-q}} \right) + \text{const.}
\end{align*}
In the replica limit (dividing through by $n$ and letting $n\rightarrow 0$) this is equal to $\logg{\rho(r,q)} =(p/2)(\log(1-q)+(q-\cos^2(\pi\,r))/(1-q)$.

To complete the calculation we find the saddle point values for $\logg{\rho(r,q)} + \mu(r,q)$ with respect to $r$ and $q$.  The saddle-point conditions are that $q=\cos(\pi\,r)$ and
\begin{align*}
    q = \frac{m}{p\,\pi} \int \frac{\e{-\frac{(1+q)t^2}{2\,(1-q)}}}{\Phi\!\left(\sqrt{\frac{q}{1-q}} \, t\right)} \,\frac{\dd t}{\sqrt{2\,\pi}}.
\end{align*}
Asymptotically this leads to an expected Gibbs' risk of
\begin{align*}
    \bar{R} \sim \frac{0.625\,p}{m}.
\end{align*}
The corrections in the asymptotic behaviour due to the fluctuations in the training ratio, $p(r|\data_m)$, compared to the annealed approximation is a factor of approximately 0.625.  The asymptotically calculation of the Gibbs' risk was a \textit{tour de force} in applying replica techniques.  It highlighted the importance of corrections due to the fluctuations in the training ratio $p(r|\data_m)$ in improving the generalisation behaviour.  A note of caution is necessary though.  The Gibbs's risk is only a good approximation to the risk that we experience when the two assumptions that we have made are satisfied.  These are that the learning algorithm finds a weight vector in $\Werm{m}$ and that the weight vector it finds is an unbiased sample from this set. This is not always true.  The perceptron learning algorithm will find a weight vector in $\Werm{m}$, but it is slightly biased.  To get empirical results close to the Gardener results it is necessary to add a randomisation element to the perceptron algorithm \citep{Berg1998}.

The Gardner calculation is widely believed to be exact, in part because it agrees with experiments and in part because it follows a tradition of calculations that have been shown to work in many different cases.  However, some of the mathematical manipulations, particularly in taking the replica limit, are difficult to prove rigorously.  In the calculation we presented we assumed replica symmetry.  In \cite{Gardner_1988} the saddle-point calculation was shown to be stable to replica symmetry breaking providing stronger justification for this step.  In many situation (e.g. when we have an unrealisable model so that $\Werm{m}$ is fragmented) replica symmetry may be broken and the calculation of the Gibbs' risk is more complicated.
We should also point out that the results are only exact in the limit $p,\,m\rightarrow \infty$ in such a way that the ratio $m/p$ remains finite.  This is necessary to ensure that the central limit theorem holds exactly and that the saddle-point evaluation of the integrals are exact.  In practice the central limit theorem and saddle-point evaluation of integrals are extremely good approximation even when $p$ and $m$ are of the order of 100.  Consequently the Gardner result are empirically found to describes quite small networks reasonable well.

Although the problem analysed here, what we have called the \textit{realisable perceptron}, is a classic problem in the literature, we believe that it is rather misleading for understanding generalisation in most machine learning applications.  The reason for this is that the dividing plane is arbitrarily chosen and does not reflect the distribution of the features, $\bm{x}$, which are spherically symmetric.  This makes this a difficult problem to learn; requiring the number of training examples, $m$, to be of order of the number of features, $p$, before we get good generalisation.  As we saw in Section~\ref{sec:results} this is not necessarily true in many real world situations.  In particular, we would argue that this example gives the impression that the asymptotic behaviour provides a good guide to what happens for small training sets.  Our belief is that this is far from true.  Most machine learning is about distinguishing classes that are fundamentally different and this is reflected in the distribution of the data, $\gamma(\bm{x},y)$.  As a result finding a good approximation is relatively easy.  That is, a relatively large proportion of weight space corresponds to rules with good generalisation performance.  Practical machine learning is about finding these good approximations rather than finding the absolute best risk machine.

\section{The asymptotic convergence of the generalisation}
\label{sec:lossland}

The asymptotic convergence of the generalisation is determined by the behaviour of $\rho(r)$ around $R_{min}$.  Here we argue that in the limit $r\rightarrow R_{min}^{+}$ that $\rho(r)\approx (r-R_{min})^a$ where the exponent $a$ is determined by the geometry of the loss landscape.  Direct measurements of these properties are unfortunately extremely challenging due to the massive size of parameter space of many modern machine learning models.  We consider the growth in the risk around the set of minimum risk weight vectors $\bm{w}^* \in \mathcal{W}_{min}$.  Note that these are not the set of weights that minimise the empirical risk $\Werm{m}$, but rather the much less tangible set of weights that would minimise the true risk.  Due to symmetries (e.g. scaling of the weights into and from a filter or node) we expect that there is a manifold of optimal weights.  Furthermore, there may be many discrete symmetries due, for example, to swapping the weights between different feature maps in CNNs or different neurons in the dense layer.  The generalisation performance depends only on the relative proportion of risks $\rho(r_1)/\rho(r_2)$ for low risk machines so these symmetries do not contribute to the growth exponent $a$ (although the existence of the continuous symmetries reduces the directions in which the risk changes which does affect the exponent $a$).  We consider one representative weight vector $\bm{w}^*\in\mathcal{W}_{min}$ and consider the subspace orthogonal to the symmetric directions.

First we consider directions where the risk increases linearly away from $\bm{w}^*$.  This may occur because the weights lie on a constraint or because the minimum lies at a discontinuity in the derivative.  In these directions the risk increases as
\begin{align*}
    R(\bm{w}) = R_{min} + (\bm{w}-\bm{w}^*)^\tr \bm{g} + O\left((\bm{w}-\bm{w}^*)^2\right)
\end{align*}
where because we are at a minimum $\bm{w}$ can only vary in a direction that is positively correlated with $\bm{g}$.  Let $d_1$ be the dimensionality of the subspace with a linear growth in the risk.  We choose to parameterise our weight space so that $\bm{w}$ represents the $d_1$ components in this space and $w_i\,g_i>0$.  Then we can approximate the distribution of risks due to this subspace where the risk grows linearly as
\begin{align*}
    \rho_l(r) \propto \int \pred{(\bm{w}-\bm{w}^*)^\tr\bm{g}>0} \dirac{R_{min} + (\bm{w}-\bm{w}^*)^\tr \bm{g}-r} \dd \bm{w}
\end{align*}
(we ignore the higher-order terms in the Taylor expansion of the risk as they will not affect the exponent $a$).  Taking the Fourier transform
\begin{align*}
    \tilde{\rho}_l(\omega) &= \int_{-\infty}^\infty \e{\ii \omega \, r} \, \rho_l(r) \, \dd r
    \propto \int \pred{(\bm{w}-\bm{w}^*)^\tr\bm{g}>0} \e{\ii\,\omega\,(R_{\min} + (\bm{w}-\bm{w}^*)^\tr \bm{g})} \, \dd \bm{w}
    \\
    &\propto \e{\ii\, \omega\,R_{min}} \prod_{i=1}^{d_1} \int_{w^*_i}^\infty \frac{1}{\omega^{d_1}} \e{\ii g_i (w_i -w_i^*)} \dd w_i = \frac{\e{\ii\,\omega\,R_{min}}}{\omega^{d_1} \prod_i g_i}.
\end{align*}
Taking the inverse Fourier transform
\begin{align*}
    \rho_l(r) = \int_{-\infty}^\infty \e{-\ii\,\omega\,r} \, \tilde{\rho}_l(\omega)
    \frac{\dd \omega}{2\,\pi} \propto \frac{1}{2\,\pi \prod_i g_i}
    \int_{-\infty}^\infty \e{\ii\,\omega\,(R_{min}-r)} \, \dd \omega 
    = \frac{(r-R_{min})^{d_1-1}}{\Gamma(d_1)\prod\limits_{i=1}^{d_1} g_i}.
\end{align*}
The constants are irrelevant to the generalisation performance.  What is relevant to the exponent $a$ is that we get a factor of $r-R_{min}$ for every direction where the risk grows linearly.

For analytic functions without constraints we would expect a minimum will, with over-whelming probability, be quadratic.  We consider the case where we have a subspace of size $d_2$ where the risks grow quadratically
\begin{align*}
    R(\bm{w}) \approx R(\bm{w}^*) + \frac{1}{2} (\bm{w}-\bm{w}^*)^\tr \mat{H} (\bm{w}-\bm{w}^*) + \cdots
\end{align*}
where $\mat{H}$ is the Hessian define in this subspace.  We ignore higher-order corrections as they don't contribute to the exponent $a$.  Following the calculation we performed for a linear minimum we can compute the Fourier transform of the distribution of risks
\begin{align*}
    \tilde{\rho}_q(\omega) &\propto \e{\ii \, \omega \,  R_{min}}
    \int_{-\infty}^\infty \e{\frac{\ii\,\omega}{2} (\bm{w}-\bm{w}^*)^\tr \mat{H} (\bm{w}-\bm{w}^*)} \dd \bm{w} = \left(\frac{2\,\pi}{\omega}\right)^{d_2/2} \, \frac{1}{\sqrt{|\mat{H}|}}.
\end{align*}
Taking the inverse Fourier transform
\begin{align*}
    \rho_q(r) \propto \frac{(2\,\pi)^{d_2/2}}{\Gamma(d_2)\,\sqrt{|\mat{H}|}} (r-R_{min})^{d_2/2-1}.
\end{align*}
Again the constants are irrelevant to the exponent $a$.  If we have both linear and quadratic terms then
\begin{align*}
    \tilde{\rho}(\omega) \propto \omega^{-d_1-d_2/2}
\end{align*}
and $\rho(r) \propto (r-R_{min})^{d_1+d_2/2-1}$ for small values of $r-R_{min}$.

We speculate that this growth exponent $a = d_1+d_2/2 -1$ is closely related to other measures of the capacity of a network.  It depends primarily on the architecture of the network rather than distribution of the data, $\gamma(\bm{x},y)$.  However, although this does tells us about the convergence towards obtaining the absolute best machine, it tells us little about the time it takes to obtain a good approximation except in a few cases such as the realisable perceptron.

\section{Risk for the unrealisable perceptron}
\label{sec:nomalclasses}

\begin{figure}[htbp]
\centering
\includegraphics[width=0.5\textwidth]{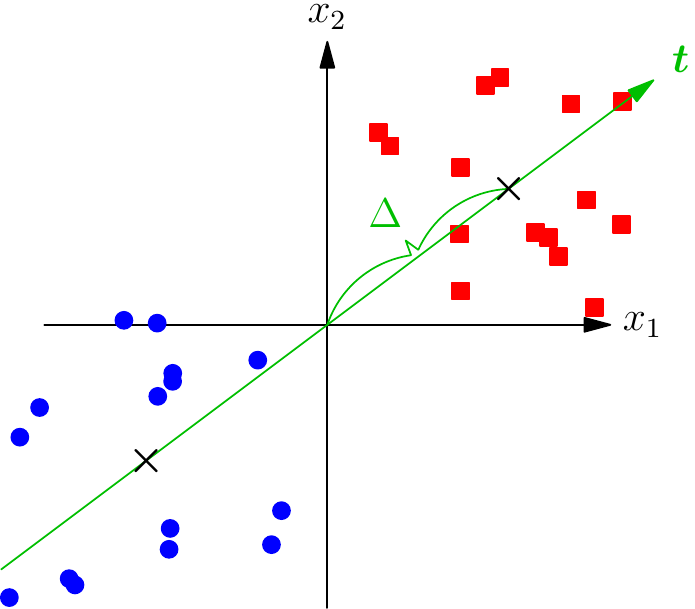}
\caption{\label{fig:orgac7ea35}
Example of a data set of $m=30$ data points drawn independently from $\gamma_{\bm{x},y}(\bm{x},y)$ with $\Delta=0.8$ and $p=2$.}
\end{figure}

We consider a very simple two-class problem where the data points $(\bm{x},y)$ are distributed according to $\gamma_{\bm{x},y}(\bm{x},y) =
\gamma_{\bm{x} \mid y}(\bm{x} \mid y)\,\Prob[y]{y=y}$
\begin{align*}
\gamma_{\bm{x}|y}(\bm{x}\mid y) &= 
 \mathcal{N}\!\left( \bm{x} \big| y\,\Delta\,\bm{t}, \mat{I}_p \right)
&
\Prob[y]{y=y} = \frac{1}{2} \left( \pred{y=1} + \pred{y=-1} \right)
\end{align*}
where $|\bm{t}|=1$. We assume that we have a training data set, $\mathcal{D}$, consisting of $m$ examples drawn independently from $\gamma_{\bm{x},y}(\bm{x},y)$.
The parameter $\Delta$ determines the separation between the means of the two classes.
We illustrate this in Fig.~\ref{fig:orgac7ea35} for the case when $p=2$ and $m=30$.

The boundary of the Bayes optimal classifier corresponds to a hyperplane orthogonal to $\bm{t}$.  We consider learning a perceptron defined by the unit length weight vector $\bm{w}$.  We call this the \textit{unrealisable perceptron} to distinguish it from the realisable perceptron discussed in Section~\ref{sec:gardner}.  Although, one should be suspicious of all simple models, be believe that this captures a component of real world machine learning problems missing from the realisable perceptron.  That is, the class structure is reflected in the distribution of the features, $\bm{x}$. As we will see a consequence of this is that the problem is easily approximated (i.e. there a many weights $\bm{w}$ which although not that close to the optimum weights, $\bm{w}^*=\bm{t}$, have very low risk).  This makes it possible to approximate with far fewer training examples than was the case for the realisable perceptron.

We consider a perceptron with weight vectors $|\bm{w}| =1$ (in a $p$-dimensional space) that makes predictions $\hat{f}(\bm{x}|\bm{w}) = \mathrm{sign}(\bm{w}^{\tr}\bm{x})$. We define a loss function for a perceptron with weights $\bm{w}$ to be
\begin{align*}
L_{\bm{w}}(\bm{x},y)=\pred{\hat{f}(\bm{x}|\bm{w})\neq y}
\end{align*}
so that the risk for a weight vector $\bm{w}$ is the expected loss over the
randomly drawn data
\begin{align*}
R(\bm{w}) = \av[\bm{x},y]{L_{\bm{w}}(\hat{f}(\bm{x}|\bm{w}),y)} 
= \Prob[(\bm{x},y)\sim\gamma]{y\,\bm{w}^{\tr}\,\bm{x}\leq 0}.
\end{align*}
We can represent a random deviate, $(\bm{x},y)$ drawn from
$\gamma_{\bm{x},y}(\bm{x},y)$ as $\bm{x}=y\,\Delta\,\bm{t} + \bm{\eta}$ where $y\sim\Prob[y]{y}$ and $\bm{\eta}\sim \mathcal{N}(\bm{0},\mat{I})$.
Defining $\cos(\theta_{\bm{w}}) = \bm{w}^{\tr}\bm{t}$ (recall
$|\bm{t}|=|\bm{w}|=1$), then
\begin{align*}
\bm{w}^{\tr} \bm{x} = \Delta\,\cos(\theta_{\bm{w}}) + \epsilon
\end{align*}
where $\epsilon = \bm{\eta}^{\tr}\bm{w}_{h}$. Since $\bm{\eta}\sim
\mathcal{N}(\bm{0},\mat{I})$  then $\epsilon$ is the sum of independent
normal components so will normally distributed.  Furthermore
\begin{align*}
\av{\epsilon} &= \av{\bm{\eta}^{\tr}}\bm{w} = 0 &
\av{\epsilon^{2}} &= \bm{w}^{\tr} \, \av{\bm{\eta}\,\bm{\eta}^{\tr}} \,\bm{w} 
= \bm{w}^{\tr} \, \mat{I} \,\bm{w} = 1
\end{align*}
since $|\bm{w}|=1$.  Therefore $\epsilon\sim \mathcal{N}(0,1)$ and
\begin{align}
\label{eq:risk}
R(\bm{w}) = \Prob[\bm{x},y]{y\,\bm{w}^{\tr}\bm{x}\leq0} 
= \Prob[\epsilon,y]{y\,\epsilon\leq -\Delta\,\cos(\theta_{\bm{w}})} 
= \Phi(-\Delta\,\cos(\theta_{\bm{w}}))
\end{align}
where $\Phi(x)$ is the cumulative distribution function for a
standard normal distribution.

The distribution of angles between two vectors in a $p$-dimensional space is
\begin{align*}
   f_\Theta(\theta) = \frac{\sin^{p-2}(\theta)}{B(\tfrac{1}{2}, \tfrac{p-1}{2})}.
\end{align*}
The distribution of risks is given by
$f_R(r) = f_\Theta(\theta(r))/\tfrac{\dd r}{\dd \theta}$ where
$r = \Phi(-\Delta\,\cos(\theta))$ or
$\theta(r) = \cos^{-1}(\Phi^{-1}(r)/\Delta)$.  Noting that
\begin{align*}
  \frac{\dd r}{\dd \theta}
  = \Delta\, \sin(\theta) \, \frac{\e{-\Delta^2\,\cos^2(\theta)/2}}{\sqrt{2\,\pi}}
\end{align*}
and writing
\begin{align*}
    \sin^{p-3}(\theta) =  \left( 1-\cos^2(\theta) \right)^{\frac{p-3}{2}} = \left( 1-\left(\frac{\Phi^{-1}(r)}{\Delta}\right)^2 \right)^{\frac{p-3}{2}}
\end{align*}
we get
\begin{align*}
    \rho(r) = \frac{\sqrt{2\,\pi}}{\Delta\, B(\tfrac{1}{2},
  \tfrac{p-1}{2})} \,
  \left( 1-\left(\frac{\Phi^{-1}(r)}{\Delta}\right)^2
  \right)^{\frac{p-3}{2}}
  \, \e{(\Phi^{-1}(r))^2/2}.
\end{align*}

The risk entropy per feature in the limit of large $p$ is thus
\begin{align*}
    \lim_{p\rightarrow\infty} \frac{s(r)}{p} =  \frac{1}{2} \logg{1-\left(\frac{\Phi^{-1}(r)}{\Delta}\right)^2}.
\end{align*}
This is illustrated in Fig~\ref{fig:entropyCurve}(a) in the main text.

If we Taylor expand $\rho(r)$ around $R_{min}$ we find for large $p$ that $\rho(r) \propto (r-R_{min})^{p/2}$ (as we would expect from Section~\ref{sec:lossland}).  If we compute the Gibbs' risk (at least in the annealed approximation) the asymptotic convergence to $R_{min}$ falls off as $p/(2m)$, which is determined by the dimensionality of the features $p$ (or the capacity of the perceptron).  This would seem to suggest that the converge to $R_{min}$ is independent on the data distribution (although clearly $R_{min}$ depends on this distribution).  However, $\Phi^{-1}(r)$ has a singularity at $r=0$, so for small $R_{min}$ the derivative $s'(r)$ that determines the Gibbs' risk is extremely rapidly changing.  Thus we only see the $p/(2m)$ convergence very late on.  To see this explicitly we can use the asymptotic approximation $\Phi^{-1}(r)\approx \sqrt{-2\log(r)}$ so that
\begin{align*}
    s(r) \approx \frac{p}{2} \logg{\frac{2}{\Delta^2} \logg{\frac{r}{R_{\min}}}}
\end{align*}
and $s'(r) = p/(2\,r\,\log(r/R_{min}))$.  Now $s'(r)$ determines the position of the Gibbs' risk. Expanding $r$ around $R_{min}$
\begin{align*}
    s'(r) \approx \frac{p}{2\,(r-R_{min})} \frac{R_{min}}{r}
\end{align*}
so that when $r-R_{min}\ll R_{min}$ we find the asymptotic behaviour predicted by the Taylor expansion, however when $r-R_{min}$ is or order $R_{min}$ we get large corrections to the asymptotic behaviours.  This rapid change in the gradient means that the actual Gibbs' risk we see will be very different to that predicated by the asymptotic behaviour and in fact it depends heavily on the distribution of data.  

\section{Hebbian learning}
\label{sec:hebb}

Computing the Gibbs' risk for the unrealisable perceptron is non-trivial because we need to compute the fluctuation in the training ratio $p(r|\data_m)$.  Rather than go through a full replica calculation instead we consider an alternative learning algorithm where we choose our weight vector $\bm{w}\propto\sum_{k=1}^m y_k\,\bm{x}_k$.  This is known in the statistical physics literature as ``Hebbian Learning''.  Here we can compute the generalisation exactly (although we make a benign approximation to obtain a closed form solution).  For this problem Hebbian learning converges somewhat faster than Gibbs' learning. Unfortunately Hebbian learning is not generally applicable and can only be used for relatively simple learning architectures.

We observe that the root-mean-squared size of the training examples, $\sqrt{\av{\|\bm{x}\|^{2}}}$, is equal to $\sqrt{\Delta^2 + p}$, so when the number of features, $p$, become large the typical distance between training examples is far larger that the distance separating the means of the two classes ($2\,\Delta$).  It is important to realise that, although in this problem, and we would argue more generally in real world problems, there is a clear distinction between the classes that is reflected in the features, nevertheless for high dimensional features this distinction is usually obscured by the natural fluctuations in the data that are not class related.

We assume are are given $m$ training examples
$\{(\bm{x}_k,y_k)| k=1,2,\ldots,m\}$ drawn
independently from $\gamma_{\bm{x},y}$.  We consider a \emph{Hebb classifier}
where we choose a weight vector $\bm{w} = \bar{\bm{w}}/|\bar{\bm{w}}|$
where
\begin{align}
\label{eq:classifier}
\bar{\bm{w}} = \sum_{k=1}^m y_k\,\bm{x}_k.
\end{align}
We can reparameterising the training examples as $\bm{x}_k =
y\,\Delta\,\bm{t} + \bm{\eta}_k$ where $\bm{\eta}_k
\sim \mathcal{N}(\bm{0},\mat{I}_p)$ (note that $\av{\bm{x}_k|y_k} = y_k\,\Delta\,\bm{t}$.  Then we can write
\begin{align*}
\bar{\bm{w}} = m\, \Delta \, \bm{w}^{*} 
+ \sum_{k=1}^m y_k\,\bm{\eta}_k.
\end{align*}
As the second term is a sum of normally distributed variables it will
itself be normally distributed with mean 0 and covariance
$m\,\mat{I}_{p}$.  Defining $\bm{z} = \tfrac{1}{m} \sum_{k}
y_k\,\bm{\eta}_k$ then (i.e. a perceptron with weights
$\bm{w}_{h}$) 
\begin{align*}
\bar{\bm{w}} = m \, \Delta \, \bm{w}^{*} + \sqrt{m} \, \bm{z}
\end{align*}
where $\bm{z}\sim \mathcal{N}(\bm{0},\mat{I}_p)$.

The cosine similarity between $\bm{w}$ and the target $\bm{t}$ is given by
\begin{align*}
\cos(\theta) = \bm{w}^{\tr}\bm{t} = \frac{m\,\Delta}{|\bar{\bm{w}}|}
\end{align*}
where
\begin{align*}
|\bar{\bm{w}}|^2 = m^2 \, \Delta^2 + m^{3/2} \bm{z}^{\tr}\bm{t}
		      + m \, |\bm{z}|^{2}.
\end{align*}
In expectation $\av{\bm{z}^{\tr}\bm{t}}=0$.  Furthermore
$|\bm{z}|^{2}$ will be chi-squared distributed with mean $p$.  For
large $p$ the chi-squared distribution will be sharply peaked
around its mean value.  As this is the limit where we are interested,
we will approximate $|\bar{\bm{w}}|$ by its RMS value of
$\sqrt{m^2\,\Delta^{2} + m\,p}$ (although we can compute the exact
risk numerically, this approximation is sufficient).  Using this
approximation
\begin{align*}
\cos(\theta) = \frac{\bar{\bm{w}}^\tr \bm{t}}{\|\bar{\bm{w}}\|} \approx \frac{m\,\Delta}{\sqrt{m^2\,\Delta^{2} + m\,p}} =
\frac{1}{\sqrt{1 + p/(m\,\Delta^{2})}}.
\end{align*}
Recalling that the risk of a weight vector with an
angle of $\theta$ from $\bm{t}$ is $\Phi(-\Delta\,\cos(\theta))$
then the expected risk using the Hebb rule is
\begin{align}
\label{eq:expectedRisk}
\bar{R} \approx \bar{R}_{approx} = \Phi \!\left( 
-\frac{\Delta}{\sqrt{1+\frac{p}{m\,\Delta^{2}}}} \right)
\end{align}
In Figure \ref{fig:entropyCurve}(b) on page~\pageref{fig:entropyCurve} in the main text we show the expected risk versus the number of training examples for feature vectors of length $p=50, 100$ and 200 and for different separations of the two distribution $\Delta=1,2,3$ and 4.  In addition, we show simulation results averaged over 100 different training runs.  Clearly, the approximation of replacing the expectation over a chi-squared distributed random variable its expected value provides a very good fit to the observed behaviour.

When $m$ is large we find from expanding Eq.~\eqref{eq:expectedRisk}, that
\begin{align*}
\bar{R} &\approx \Phi(-\Delta) \, \left( 1 + \frac{p}{2\,m} \right),
&
\logg{\bar{R}} &\approx \logg{\Phi(-\Delta)} + \frac{p}{2\,m}.
\end{align*}
Here we have used the approximation that $\Phi(-\Delta) \approx
\e{-\Delta^{2}/2}/(2\,\pi\,\Delta)$ which is a good approximation as
$\Delta$ becomes moderately large.  Note that $R_{min} = \Phi(-\Delta)$ so that $\bar{R} - R_{min} = R_{min}\, p/(2\,m)$.  This is a dramatically faster convergence rate than we might expect from extrapolation of the asymptotic result (by a factor of $R_{min}$).

Clearly the realisable and unrealisable perceptron behave very differently. In particular, the unrealisable perceptron is far easier to approximate than the realisable perceptron because we are learning a class distinction that is reflected in the features, $\bm{x}$.  Although, the convergence still depends on the dimensionality, $p$, of the feature vectors, the separation parameter of the classes, $\Delta$, dramatically speeds up the convergence (for Hebbian learning by a factor of $R_{min}=\Phi(-\Delta)$).  The growth in the distribution of risks $\rho(r) \approx (r-R_{min})^a$ around $R_{min}$ is misleading in understanding the initial reduction in the risk.  Intuitively it is clear that we can obtain a good approximation for the unrealisable perceptron problem even when $\bm{w}^\tr\bm{t} = \cos(\theta)$ is considerably less than 1.  On the other hand if we concern ourselves with reaching $R_{min}$ then we require $\bm{w}^\tr\bm{t}$ to be very close to 1. It is therefore not surprising that asymptotically this convergence behaviour is the same as that we would experience for the realisable perceptron.  Because much of statistical learning theory is focused on asymptotic convergence, we believe that it is largely uninformative about the rapid \textit{approximability} of many problems in machine learning.

\section{MCMC details}
\label{sec:mcmcdetails}

MCMC can be very time consuming with the equilibration time often growing exponentially with $\beta$.  As we have shown empirically as we increase $\beta$ we decrease the Boltzmann risk, but $\rho(r)$ decreases exponentially with $r$ (often with a very large exponent) so it becomes much less likely to move to a low risk weight if we make large jumps.  This would not, on its own, prevent us from sampling $\rho(r)$ at low $r$ using MCMC, however when the risk landscape $R(\bm{w})$ is rugged so that we have to overcome risk (or low entropy) barriers then we would expect the equilibration times are likely to grow exponentially.  Because of the difficulty of knowing if MCMC has equilibrated we have to be cautious in interpreting our results.  In particularly they may not be reliable for large $\beta$ particularly for \cifarstar{} where finding any approximating model is very difficult.

Most of the empirical work described here is relatively recent and we are still working on developing efficient methods for performing MCMC.  Although computing the Boltzmann risk is of academic rather than practical interest, nevertheless we believe that this opens a new window into understanding generalisation in complex machine learning models so that doing this efficiently is valuable.  One approach to speeding up MCMC is what we call the \textit{minibatch proposal MCMC}.  Starting from a weight vector $\bm{w}_0$ with risk $R(\bm{w}_0)$ we perform a series of Metropolis steps using approximate risks evaluated using minibatches to give a series of weight updates $\bm{w}_1$, $\bm{w}_2$, \ldots, $\bm{w}_n$ with approximate risks $R_{\mathcal{B}_i}(\bm{w}_i)$ computed using a randomly chosen minibatch $\mathcal{B}_i\in\data$.  The acceptance probability $A(\bm{w}_0\rightarrow \bm{w}_n)$ for going from $\bm{w}_0$ to $\bm{w}_n$ divided by the acceptance probability $A(\bm{w}_n\rightarrow \bm{w}_0)$ for going from $\bm{w}_n$ to $\bm{w}_0$ using the same set of minibatches at each step is equal to
\begin{align*}
    \frac{A(\bm{w}_0\rightarrow \bm{w}_n)}{A(\bm{w}_n
    \rightarrow \bm{w}_0)} = \e{-\beta\left(R_{\mathcal{B}_1}(\bm{w}_1)
    -R(\bm{w}_0) + \sum_{i=2}^{n} (R_{\mathcal{B}_i}(\bm{w}_i)-
    R_{\mathcal{B}_{i-1}}(\bm{w}_{i-1}))\right)}
    = \e{-\beta\left(R_{\mathcal{B}_n}(\bm{w}_n) - R(\bm{w}_0)\right)},
\end{align*}
Using the Metropolis-Hastings update equation we  decide whether to accept the proposal $\bm{w}_n$ by computing $R(\bm{w}_n)$ using the full data set $\data$ and accept the proposal if $R(\bm{w}_n) \leq R_{\mathcal{B}_n}(\bm{w}_n)$ or with a probability $\exp\!\left(\strut -\beta R(\bm{w}_n) + \beta R_{\mathcal{B}_n}(\bm{w}_n)\right)$.  This allows us to obtain decorrelated samples quicker.

An alternative that we are yet to explore fully is to choose as our proposal distribution a set of weights obtained by moving orthogonal to the gradient directions computed on a training set.  This acts like a hybrid MCMC algorithm \citep{DuaEtAl87} in that the new weights should have a close risk to that of the old weights.

\section{Experimental details}
\label{sec:experiment}

The experiments carried out on MNIST with \cnnone{} were based on a pytorch \href{https://github.com/pytorch/examples/tree/master/mnist}{example}, although we removed the dropout layers altogether.  The network consisted of the following layers:
\begin{verbatim}
Conv2d(1, 32, (3, 3), padding=1)
ReLU()
Conv2d(32, 64, (3, 3), padding=1)
ReLU()
MaxPool2d((2, 2))
Flat()
Linear(9216, 128)
ReLU()
Linear(128, 10)
Sofmax()
\end{verbatim}

We experimented with a network \cnnstar, where we put a fully connected layer before the first convolutional layer.  This was a much more expressive machine that can potentially learn the same mapping as \cnnone.  Even at $\beta=1000$ the Boltzmann risk was not significantly different from the risk of a random machine $0.9$.  Although this machine has a slightly larger number of weights it is the fact that it breaks the locality constraint of the CNN that prevents this machine from generalising.

In the second set of experiments on \cifar{} we use a slightly more powerful network \cnntwo with layers:
\begin{verbatim}
Conv2d(3, 24, (3, 3), padding=1)
ReLU()
BatchNorm2d(24)
Conv2d(24, 48, (3, 3), padding=1)
ReLU()
BatchNorm2d(48)
MaxPool2d((2, 2))
Conv2d(48, 48, (3, 3), padding=1)
ReLU()
BatchNorm2d(48)
Conv2d(48, 48, (3, 3), padding=1)
ReLU()
BatchNorm2d(48)
MaxPool2d((2, 2))
Flat()
Linear(3072, 768)
ReLU()
Linear(768, 10)
\end{verbatim}
We also considered an MLP with layers:
\begin{verbatim}
Flat()
Linear(3072, 768)
ReLU()
Linear(768, 10)
\end{verbatim}
This is identical to the the last two layers of \cnntwo.

In the problem \cifarstar{} we trained \cnntwo{} on the training set of \cifar{} but with randomised labels to obtain weights $\bm{w}^*_R$.  We then used the predictions of this "teacher network" on the \cifar{} test set as new labels (i.e. $y = \hat{f}(\bm{x}|\bm{w}^*_R)$).  We computed the Boltzmann risk for \cnntwo{} using \cifarstar.  This is a realisable model in that there exists a weight vector that will perfectly predict the teacher.  The Boltzmann risk were barely statistically different from random risks of $0.9$ and the extrapolation shown in Fig.~\ref{fig:cifar} in the main text should be treated cautiously.  Note that the derivative of the risk entropy $s'(r)$ is (in the saddle-point approximation) equal to $\beta$ so this derivative should be 1000 around $R_B(1000)\approx 0.9$.

In a rather informal study we examined the rise in the risk, $R(\bm{w})$, as we perturbed $\bm{w}$ around the weights $\bm{w}^*_R$ and also a set of weights $\bm{w}^*_C$ obtained by training \cnntwo{} on the real \cifar{} labels.  Surprisingly perhaps $\hat{f}(\bm{x}|\bm{w}^*_R+ \bm{\Delta})$ appears to have more stable outputs than  $\hat{f}(\bm{x}|\bm{w}^*_C+ \bm{\Delta})$ to small perturbations $\bm{\Delta}$.  This may reflect the fact that more of the nodes where truncated to 0 by the ReLU layer around $\bm{w}^*_R$ than around $\bm{w}_C^*$.  Unfortunately it is difficult to measure the growth exponent $a$  in $\rho(r)\sim(r-R_{min})^a$ empirically as we can only estimate the risk $R(\bm{w})$ from a finite data set $\mathcal{D}$ which is insensitive to small changes of risks.  Our observation on the stability of $\hat{f}(\bm{x}|\bm{w})$ around optimal values of $\bm{w}$ lends some support to the notion that the behaviour of $\rho(r)$ around $R_{min}$ might not be very informative about the generalisation behaviour when we only have a moderate number of training examples.

\end{document}